\documentclass{article}

\usepackage[nonatbib]{arxiv}

\usepackage[utf8]{inputenc} 
\usepackage[T1]{fontenc}    
\usepackage{hyperref}       
\usepackage{url}            
\usepackage{booktabs}       
\usepackage{amsfonts}       
\usepackage{nicefrac}       
\usepackage{microtype}      
\usepackage{lipsum}
\usepackage{graphicx}

\usepackage[ruled, vlined]{algorithm2e}
\usepackage{amsmath}
\usepackage{multirow}
\usepackage{bm}
\usepackage{color, colortbl}
\definecolor{Gray}{gray}{0.9}
\definecolor{LightCyan}{rgb}{0.6,1,1}
\usepackage{booktabs}

\newcommand{\fgsm}{{FGSM}}
\newcommand{\ifgsm}{{I-FGSM}}
\newcommand{\cw}{{CW}}
\newcommand{\lb}{{L-BFGS}}
\newcommand{\maxiter}{\textit{max iterations}}
\newcommand{\lr}{\textit{learning rate}}

\title{Towards Leveraging the Information of Gradients\\in Optimization-based Adversarial Attack}

\author{
  Jingyang Zhang \\
  Department of Electronic Engineering\\
  Tsinghua University\\
  Beijing, China \\
  \texttt{jingyang15@mails.tsinghua.edu.cn} \\
   \And
  Hsin-Pai Cheng \\
  ECE Department\\
  Duke University\\
  Durham, NC 27708\\
  \texttt{hc218@duke.edu} \\
   \And
  Chunpeng Wu \\
  ECE Department\\
  Duke University\\
  Durham, NC 27708\\
  \texttt{chpwu@outlook.com} \\
   \And
  Hai Li \\
  ECE Department\\
  Duke University\\
  Durham, NC 27708\\
  \texttt{hai.li@duke.edu} \\
   \And
  Yiran Chen \\
  ECE Department\\
  Duke University\\
  Durham, NC 27708\\
  \texttt{yiran.chen@duke.edu} \\
}

\begin{document}
\maketitle

\begin{abstract}
In recent years, deep neural networks demonstrated state-of-the-art performance in a large variety of tasks and therefore have been adopted in many applications. 
On the other hand, the latest studies revealed that neural networks are vulnerable to adversarial examples obtained by carefully adding small perturbation to legitimate samples.
Based upon the observation, many attack methods were proposed.
Among them, the optimization-based CW attack is the most powerful as the produced adversarial samples present much less distortion compared to other methods. 
The better attacking effect, however, comes at the cost of running more iterations and thus longer computation time to reach desirable results. 
In this work, we propose to leverage the information of gradients as a guidance during the search of adversaries.
More specifically, directly incorporating the gradients into the perturbation can be regarded as a constraint added to the optimization process. 
We intuitively and empirically prove the rationality of our method in reducing the search space. 
Our experiments show that compared to the original CW attack, the proposed method requires fewer iterations towards adversarial samples, obtaining a higher success rate and resulting in smaller $\ell_2$ distortion.
\end{abstract}


\section{Introduction}

With the rapid growth in network depth and width as well as the improvement in network structure and topology, the use of deep neural networks (DNNs) has been successfully extended to many applications. For instance, the state-of-the-art DNNs can achieve extremely high accuracy in image classification, which could be even higher than the level humans can reach. 

However, the latest studies \cite{szegedy2013intriguing} revealed the high vulnerability of neural network models to adversarial attacks: adding a carefully designed, small perturbation to an image could result in a rapid decrease in classification confidence or even misclassification of a well-trained network, even though the perturbation is too small to be distinguished by humans. 
Such images with small perturbations, namely \textit{adversarial samples}, raise a severe security threat for deep learning technology. 
Extensive research studies have been carried out in adversarial attacks to explore the vulnerability of neural networks as well as in defense techniques to protect the systems and applications. 


Among many attack methods proposed in recent years, the fast gradient sign method (FSGM) \cite{goodfellow2014explaining} and its iterative variation \cite{kurakin2016adversarial} have drawn significant attention. 
The methods tend to exploit the gradients of classification loss to craft adversarial images.
The principle is pretty straightforward as gradients are correlated to the direction, a change along which potentially influences the classification in the most significant way.  
The methods, however, treat every pixel the same, that is, the magnitude of introduced perturbation is exactly the same for all the pixels. 
This often makes the adversarial samples more noticeable since humans are much more sensitive to the difference in low variance area \cite{liu2010just}, e.g., the background of an image. 
Compared to the original images, the adversaries generated by \fgsm{} could be a lot blurred.
Examples include Figure 2(d) in \cite{goodfellow2014explaining} and Figure 4 in \cite{kurakin2016adversarialphysical}.


CW attack \cite{carlini2016towards} is another widely adopted adversarial attack method. 
As an optimization-based approach, CW attack first defines an objective or a loss function,
and then searches for the optimal perturbation while maintaining the distortion small during the procedure.
As such, \cw{} could produce adversaries with much smaller and more imperceptible distortion. 
The performance, however, comes at the cost of speed. 
For instance, \cite{carlini2016towards} performed 200,000 optimization iterations for every image when evaluating their method.
Such a high computation cost is not practical, especially considering the fast growth in dataset size. 
In addition, the implementation of CW attack can sometimes be pretty tricky and requires careful parameter selection to obtain the desirable data reported in the paper. 
This is indeed a common scenario in optimization based methods. 



\textit{Our work aims at a strong and effective adversarial attack.} 
We propose to leverage the gradient information in the search of adversarial examples. 
Inspired by the FGSM concept, the gradients are used as a guidance in calculating the perturbation.  
Unlike FGSM that utilizes a universal magnitude to all the pixels, however, our approach will assign each pixel with its own magnitude of perturbation. 
The magnitude optimization uses the same loss function as CW, which corresponds to the general expectation in crafting adversarial attacks. 
In a nutshell, our method constrains the perturbation with the gradients, which in turn reduces the search space of adversarial examples.

In the work, we intuitively prove the rationality of our method in reducing the search space. 
Empirically, our experimental results show that our method can reach a higher attack successful rate while applying smaller distortion and requiring much fewer iterations than the original CW attack method \cite{carlini2016towards}. The effectiveness of our method is further demonstrated through the comparison with I-FGSM  \cite{kurakin2016adversarial} and L-BFGS  \cite{szegedy2013intriguing}.



\section{The Proposed Method}

\begin{algorithm}
\KwIn{The legitimate image $\boldsymbol{x}$, the target class $t$, the initial magnitude $\theta_0$.}
\KwOut{The adversarial image $\boldsymbol{x}'$.}
$\boldsymbol{x}_0'\gets \boldsymbol{x},~i\gets 0$\\
\While{$C(\boldsymbol{x}_i')\ne t$ and $i<out\_step$}{
    $\boldsymbol{\theta}\gets \theta_0,~ \boldsymbol{g}\gets\nabla_{\boldsymbol{x}_i'}f(\boldsymbol{x}'_i,t)$\\
    $\boldsymbol{x}_{i+1}'\gets\boldsymbol{x}_{i}'-\boldsymbol{\theta}\cdot\boldsymbol{g}/||\boldsymbol{g}||_2$\\
    loss$~\gets||\boldsymbol{x}_{i+1}'-\boldsymbol{x}||_2^2+c\cdot f(\boldsymbol{x}_{i+1}',t)$\\
    minimize loss and update $\boldsymbol{\theta}$\\
    $i\gets i+1$\\}
\Return $\boldsymbol{x}'\gets\boldsymbol{x}_i'$
\caption{Our algorithm to leverage gradients $\boldsymbol{g}$ in the search of adversarial examples.}
\label{algorithm}
\end{algorithm}

\begin{figure}
\centering
\includegraphics[width=0.9\textwidth]{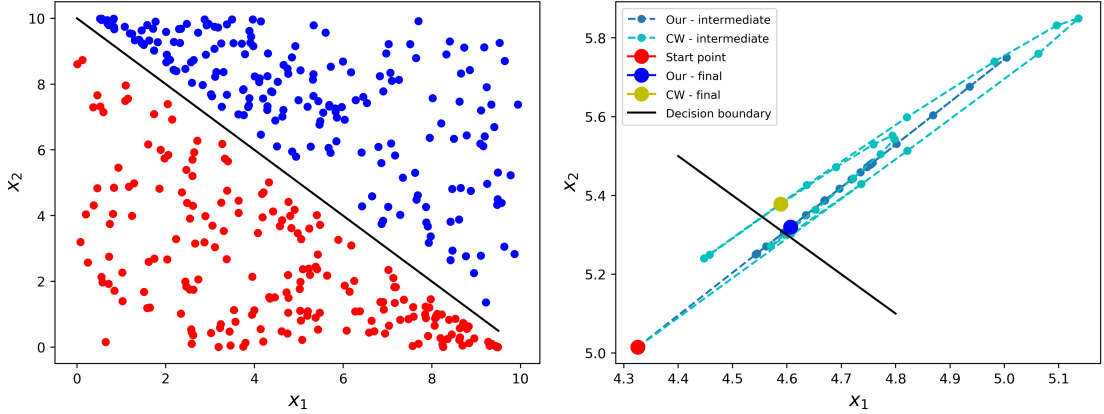}
\caption{The left figure shows the testing points of our synthetic dataset. The right figure shows the intermediate and final result of our and \cw{} attack. One can see that the \cw's final steps and ours' steps nearly coincide, indicating that we could use gradients to reduce search space. Note that the decision boundary shown here is an approximate one rather than the real one learned by the classifier.}
\label{intuition}
\end{figure}

As aforementioned, although \fgsm{} and \ifgsm{} can quickly use gradients to generate the perturbations, the large magnitude of the universal perturbations applied on each pixel could make such perturbations easily perceptible to human eyes. \cw{} attack, which optimizes the perturbation generation process using a well-defined loss function, can solve this issue by exploring the adversarial images with small perturbations. 
However, the computational cost of \cw{} attack can be huge since a large number of iterations need to be performed to minimize the loss and adjust the constant $c$. 
This fact greatly hinders applications of  such optimization-based attacks in processing a large dataset.

Thus, we propose to incorporate gradient information in the optimization process of \cw{} attack to guide the search process and reduce the search time. 
We use $\boldsymbol{\delta}$ to denote the perturbation, i.e., $\boldsymbol{x}'=\boldsymbol{x}+\boldsymbol{\delta}$. While \cw{} updates $\boldsymbol{\delta}$ during each iteration, we set $\boldsymbol{\delta}=-\boldsymbol{\theta}\cdot\boldsymbol{g}/||\boldsymbol{g}||_2$ (the multiplication here is element-wise), and use Adam to update $\boldsymbol{\theta}$ instead. Each element of $\boldsymbol{\theta}$ controls the perturbation magnitude for the corresponding pixel, and we initialize all elements in $\boldsymbol{\theta}$ with $\theta_0$. $\boldsymbol{g}$ is the gradient of hinge loss $f(\boldsymbol{x},t)=\max(\max\{Z(\boldsymbol{x})_i:i\ne t\}-Z(\boldsymbol{x})_t,-\kappa)$ w.r.t $\boldsymbol{x}$. This hinge loss corresponds to the goal of causing the network to classify the input as a target class, so its gradients could tell us how we should perturb the legitimate image. Normalized gradient $\boldsymbol{g}/||\boldsymbol{g}||_2$ is taken here so that the norm of $\boldsymbol{\delta}$ is independent from the scale of $\boldsymbol{g}$. Note that although $\boldsymbol{\theta}\cdot\boldsymbol{g}/||\boldsymbol{g}||_2$ will not strictly follow the direction of $\boldsymbol{g}/||\boldsymbol{g}||_2$ or move along that straight line like the way $\epsilon\cdot\boldsymbol{g}/||\boldsymbol{g}||_2$ does, this technique still works as shown by experimental results. The pseudocode is shown in Algorithm \ref{algorithm}. The $out\_step$ is designed for high-confidence adversaries, which are more likely to transfer to another model, since they usually needs greater perturbation.

\subsection{Intuitive Explanation}
We first illustrate the principle of our method in a simple scenario. We create a synthetic dataset which consists of data points lying on the 2D plane. As shown in Figure \ref{intuition}, the dataset has two classes. Each of them has 1,000 points for training and 200 for testing. It is a naive classification problem whose ideal decision boundary is $x_1+x_2=10$, where $(x_1,x_2)$ represents each point.

We train a small multi-layer perceptron as the classifier, which reaches $100\%$ accuracy on both training and testing data. We then apply CW and our method to find the adversary for a randomly picked point. The intermediate and final result of the search process is presented in Figure \ref{intuition}. By constraining the perturbation with gradients, our method actually reduces the search space. In this simple scenario where the decision boundary is a straight line, the fastest way to find an adversary is to move along the direction of the gradient, which is perpendicular to the decision boundary. That is exactly what our method's search trajectory looks like. Also note that the last several steps of CW attack directly fall into the search space of ours. The fact indicates the reduction of search space resulted by our method doesn't exclude the optimal results CW attack tries to find, which in turn proves the rationality of our method. Even though this is too simple an example for the much more complex image recognition tasks and neural networks, it still offers some intuitive thoughts. The effectiveness of our method will be further demonstrated by experimental results.

\section{Experimental Evaluations}

\noindent \textbf{Dataset}. We perform our experiments on CIFAR-10 \cite{krizhevsky2014cifar} and ImageNet \cite{imagenet_cvpr09}. For CIFAR-10, we pick the first 1,000 test images when applying attacks. ImageNet is a large-scale dataset which has 1,000 classes, and we randomly choose 500 validation images from 500 different classes to evaluate our method. The target label of the attack is also randomly chosen while being ensured to vary from the true label.

\textbf{Model Topology}. For CIFAR-10, we choose the same CNN used in \cite{carlini2016towards}, and it reaches the accuracy of $76.7\%$ on test images. For ImageNet, we use the pre-trained Inception v3 from Tensorflow \cite{tensorflow2015-whitepaper}.


\textbf{Baseline}. We compare our method with \cw{} attack, \lb, and \ifgsm. We use the $\ell_2$ version of \ifgsm{} to ensure comparisons are made under the same distance metric. Also the classification loss for \ifgsm{} and \lb{} is changed to the aforementioned hinge loss for direct comparison. We adopt the implementation in Cleverhans \cite{papernot2018cleverhans} for these baselines.

\textbf{Parameter Setting}. For \ifgsm, we gradually increase the maximum allowed perturbation magnitude $\epsilon$ and stop when it successfully reaches adversaries for all test images. For \lb{} and \cw, we first fix \maxiter{} to 100, which specifies the maximum number of iterations the attack can update its perturbation under a certain constant $c$. Then we tune their parameters, e.g. the \lr{} for CW attack, to make them first reach high attack success rate within relative few iterations, since the attack's success should be of the highest priority. This tuning process is based either on the empirical results reported in original papers or on our own experimental observations. When implementing our own method, we empirically assign values to the parameters, also trying to guarantee successful attacks with only a few iterations.


\subsection{Result Analysis}
When running experiments, we mainly focus on two aspects: the attack success rate and the $\ell_2$ distance. An adversary is called a success if the neural network indeed classifies it as the target class. The average $\ell_2$ distance between the legitimate image and its adversarial counterpart is calculated to show how much distortion the algorithm introduced. When the success rate is not 100$\%$, we only take the average of successes. We also pay great attention to the average total number of update iterations of \cw{} and our method. For both methods, each iteration corresponds to a single step made by Adam optimizer. We also apply the same \textit{abort early} technique in the implementation of our method as \cw, i.e., when no improvement is gained, both algorithms will abort early to avoid the meaningless search. Thus, the number of iterations could indicate the amount of the computation taken by the algorithms. However, since \ifgsm{} just computes the gradient every iteration without any optimization, and \lb{} performs line search within each of its so-called iteration, the computational cost behind each step is not comparable to our method or CW attack. Therefore, there is no point or is difficult to conduct a fair comparison between them in this aspect.

\begin{table*}[!t]
  \caption{The success rate and $\ell_2$ distance for CIFAR-10. We claim success if the network indeed classifies the adversary as the target class. When the success rate is not 1, the $\ell_2$ distance is the mean of only successes. The numbers in the first row correspond to the \textit{confidence} of the resulting adversary.}
  \centering
  \begin{tabular}{lrrrrrrrrrrrrr}
    \toprule
      &\multicolumn{2}{c}{0} &\multicolumn{2}{c}{5} &\multicolumn{2}{c}{10} &\multicolumn{2}{c}{15}
      &\multicolumn{2}{c}{20}
      &\multicolumn{2}{c}{25}
      \\
      \cmidrule(lr){2-3}\cmidrule(lr){4-5}\cmidrule(lr){6-7}\cmidrule(lr){8-9}\cmidrule(lr){10-11}\cmidrule(lr){12-13}
     &prob &dist &prob &dist &prob &dist
     &prob &dist &prob &dist &prob &dist \\
    \midrule \midrule
    
    \textbf{Our1} 
    &\textbf{100}$\%$ &\textbf{0.856}  
    &\textbf{100}$\%$ &\textbf{1.008}  
    &\textbf{100}$\%$ &\textbf{1.146}  
    &\textbf{100}$\%$ &\textbf{1.280}  
    &\textbf{100}$\%$ &\textbf{1.415}
    &\textbf{100}$\%$ &\textbf{1.547}
    
    \\

    CW1  &100$\%$ &3.220 
         &100$\%$ &3.609 
         &100$\%$ &3.937 
         &100$\%$ &4.225 
         &100$\%$ &4.475
         &100$\%$ &4.708
         
         \\
         
    CW3  &100$\%$ &2.459 
         &100$\%$ &2.685 
         &100$\%$ &2.888 
         &100$\%$ &3.109 
         &100$\%$ &3.335
         &100$\%$ &3.531
         
         \\

    CW6  &100$\%$ &1.876 
         &100$\%$ &1.977 
         &100$\%$ &2.032 
         &100$\%$ &2.058 
         &100$\%$ &2.089
         &100$\%$ &2.096
         
         \\
    
    L-BFGS2 &100$\%$ &0.941 
            &100$\%$ &1.152 
            &100$\%$ &1.306 
            &100$\%$ &1.533 
            &100$\%$ &1.715
            &100$\%$ &1.885
            
            \\
    
    I-FGSM &100$\%$ &1.409 
           &100$\%$ &1.531 
           &100$\%$ &1.625 
           &100$\%$ &1.704 
           &100$\%$ &1.754
           &100$\%$ &1.794
           
           \\
      \bottomrule
  \end{tabular}
  \label{CIFAR prob dist}
\end{table*}

\begin{table}[t]
  \caption{The average iterations used for CIFAR-10. The numbers in the first row correspond to the \textit{confidence} of the resulting adversary.}
  \centering
  \begin{tabular}{lrrrrrrr}
    \toprule
      &\multicolumn{1}{c}{0} &\multicolumn{1}{c}{5} &\multicolumn{1}{c}{10} &\multicolumn{1}{c}{15}
      &\multicolumn{1}{c}{20}
      &\multicolumn{1}{c}{25}
      &\multicolumn{1}{c}{30}\\
    \midrule \midrule
    Our1  &28 &28 &30 &34 &38 &42 &47\\
    CW1  &27 &25 &23 &22 &22 &21 &21 \\
    CW3  &109 &108 &108 &106 &103 &100 &98\\
    CW6  &197 &200 &207 &213 &219 &223 &228 \\ \bottomrule
  \end{tabular}
  \label{CIFAR iter}
\end{table}

\begin{table}[t]
  \centering
  \caption{The success rate, $\ell_2$ distance, and average iterations for the adversaries of ImageNet.}
  \begin{tabular}{lrrr}
    \toprule
      &prob &dist &iter \\
    \midrule \midrule
    \textbf{Our1}  &\textbf{100}$\%$ &\textbf{1.083} &\textbf{52}\\
    CW1  &99.6$\%$ &4.978 &36 \\
    CW3  &100$\%$ &3.030 &116\\
    CW6  &100$\%$ &1.813 &216\\
    CW10 &100$\%$ &1.497 &312\\
    L-BFGS2 &95.8$\%$ &2.000 &/ \\
    I-FGSM &100$\%$ &1.625 &/\\
    \bottomrule
  \end{tabular}
  \label{ImageNet}
\end{table}

For CIFAR-10, we generate adversaries when \textit{confidence} is in the set $\{0,5,10,15,20,25,30\}$. The results are shown in Table \ref{CIFAR prob dist} and \ref{CIFAR iter}. Note that our method reaches the smallest $\ell_2$ distortion while successfully attacking all test images. While Our1 only spends $34\%$ of the iterations of CW3, it reaches the $\ell_2$ distortion which is only $41\%$ of that of CW3. Besides, even though CW6 searches for $4\times$$-$$5\times$ more iterations than Our1, its perturbation is still $66\%$ greater than ours. Compared with L-BFGS and I-FGSM, our method could still produce superior results.

For ImageNet, we only present the results of adversaries with \textit{confidence} equals to $0$. As shown in Table \ref{ImageNet}, Our1 successfully finds all adversaries and meanwhile introduces the least amount of perturbation. On the contrary, CW attack doesn't produce comparable results even with $6\times$ of our iterations. I-FGSM and L-BFGS2 also lead to inferior results than ours, meanwhile the latter one takes more computational cost. Thus, it is proved that our method maintains its effectiveness on the large-scale dataset.

\section{Conclusion}
Adversarial attack has recently drawn significant attention in the deep learning community. 
The gradient-based adversarial example crafting scheme, i.e, 
FGSM and its variants, often introduce visually perceptible perturbation on the adversarial examples.
CW attack, as an optimization-based scheme, solves the above problem by searching for the adversaries with small distortion. 
However, the incurred high computational cost could be intolerant in real applications. 
In this work, we propose to leverage the gradient information in the optimization process of crafting adversaries by including it into the perturbation part. 
We illustrate that our proposed method can reduce the search space of the adversarial examples and thus leads to fewer iterations of the search. 
Experimental results show that compared to other tested methods, our method can also achieve a higher attack efficiency and a smaller perturbation or fewer iterations.



\bibliographystyle{unsrt}
\bibliography{references.bib}

\end{document}